# Neural daylight control system

GRIF Horațiu Ștefan
"Petru Maior" University of Tg.Mureș

**ABSTRACT**

The paper describes the design, the implementation of a neural controller used in an automatic daylight control system. The automatic lighting control system (ALCS) attempt to maintain constant the illuminance at the desired level on working plane even if the daylight contribution is variable. Therefore, the daylight will represent the perturbation signal for the ALCS. The mathematical model of process is unknown. The applied structure of control need the inverse model of process. For this purpose it was used other artificial neural network (ANN) which identify the inverse model of process in an on-line manner. In fact, this ANN identify the inverse model of process + the perturbation signal. In this way the learning signal for neural controller has a better accuracy for the present application.

## 1 INTRODUCTION

Neural networks have been proved a powerful tool in intelligent control. Numerous successful applications have been found in supporting and improving the control industry.[6] Artificial neural networks have been applied very successfully in the identification and control of dynamic systems. The universal approximation capabilities of the multilayer perceptron make it a popular choice for modeling nonlinear systems and for implementing general-purpose nonlinear controllers.[2] Multi layer perceptron (MLP) networks are composed of perceptron "type" units or nodes, which are arranged into layers where the outputs of the nodes in one layer constitute the inputs to the nodes in the next layer. The signals received by the first layer are the training inputs and the network's response is the outputs of the last layer (Figure 1a). Each of the nodes has associated with it a weight vector and a transfer (or activation) function (denoted by *F*), where the dot product of the weight vector and the incoming input vector is taken, and the resultant scalar is transformed by the activation function (Figure 1b). For a suitable arrangement of nodes and layers, and for appropriate weight vectors and activation functions, it can be shown that this class of networks can reproduce any logical function exactly and can approximate any continuous nonlinear function to within an arbitrary accuracy. [1, 5]

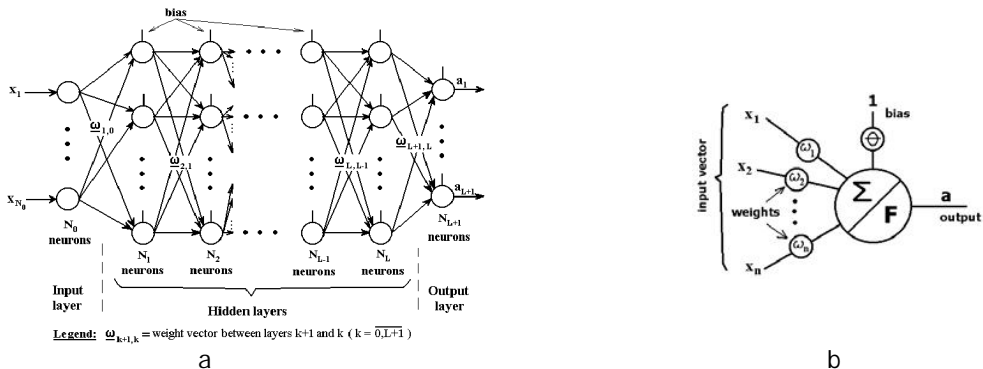

**Figure 1** Multi layer perceptron network (a) and the configuration of the perceptron (b) [5]

## 2 THE ALCS BLOCK DIAGRAM

In Figure 2 is presented the block diagram of the ALCS where, are denoted with: $E_{desired}$ – the desired illuminance on the working plane; $E_{measured}$ – the measured illuminance on working plane; $E_{real}$ – the illuminance on the working plane; $E_{daylight}$ – the daylight illuminance on working plane; $E_{electric}$ – the illuminance on working plane due to electric light; $\varepsilon$ - control error; $\Delta\varepsilon$ - change in control error; $U$ – control action (command).

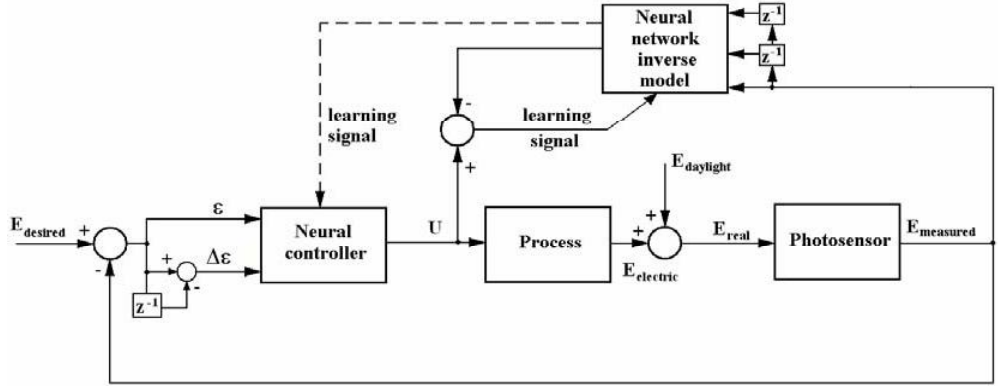

**Figure 2** Block diagram of the ALCS

The neural controller is implemented as position type. The controller, based on the values of $\varepsilon$ (control error – the difference between $E_{desired}$ and $E_{measured}$) and $\Delta\varepsilon$ (change in control error - the difference between current control error and anterior control error) will generate the control action denoted by $U$. The control action $U$ will be applied to the process, in the purpose to maintain the illuminance in working plane close to the desired illuminance $E_{desired}$. At step $k$ the ANN of controller is trained using the values of $\varepsilon(k\text{-}1)$, $\Delta\varepsilon(k\text{-}1)$ and $U_{IM}(k)$, where $U_{IM}$ is the command generate when apply to the ANN of inverse model the current and the last two values of desired illuminance. At step $k$ the ANN of inverse model is trained using the values of $U(k)$ and $E_{measured}(k)$, $E_{measured}(k\text{-}1)$, $E_{measured}(k\text{-}2)$.

## 3 EXPERIMENTAL RESULTS

The behavior of the proposed ALCS (Figure 2) was simulated using Matlab. For this purpose, the process block was implemented with a look-up table (LUT) of measured data at the input and the output of process during night condition (Figure 3). The process encapsulates a digital ballast and two 36 W fluorescent lamps [4, 5].

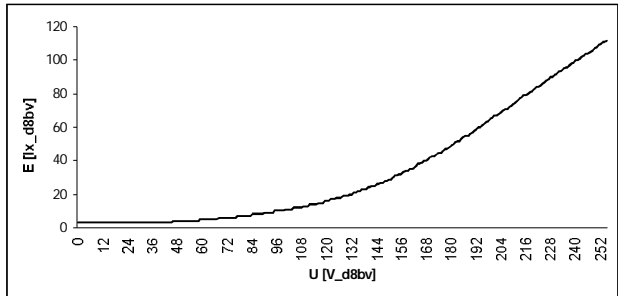

**Figure 3** The experimental model of process [4, 5]

The meaning of abbreviation *d8bv*, used in Figure 3, is "digital 8 bits value". The value 100 lx$_{d8bv}$ represents the equivalent value obtained by conversion with 8 bits A/D converter of the 500 lx, which represents the illuminance on working plane measured by an analog luxmeter. The value 127 V$_{d8bv}$ represents, by conversion with 8 bits D/A converter, the equivalent for a d.c. voltage with value 5V$_{dc}$. [5].
The controller and the inverse model was implemented with ANNs (using nnet toolbox of Matlab) with three layers (input layer, hidden layer, output layer). For controller, the ANN has two inputs and for inverse model the ANN has three inputs. Both ANNs, has in the hidden layer three neurons and in the output layer has one neuron.
    The neurons from hidden layer has the hyperbolic tangent function as activation function:

$$F(x) = \frac{e^x - e^{-x}}{e^x + e^{-x}} \qquad (1)$$

    The neuron from the output layer has linear function as activation function:
$$F(x) = x \qquad (2)$$

Both ANNs are trained on-line using the back-propagation training rule. The learning rate for both ANNs was set to $\gamma = 0,15$.

The universes of discourse of $E_{desired}$ and $E_{measured}$ are fixed to the interval of integers [0 ; 255] ($lx_{d8bv}$), due to the 8 bits A/D and D/A converters. The inputs and the output of ANNs are scaled [3], which implies the conversions of the universes of discourse of the $E_{desired}$ and $E_{measured}$ variables in the intervals [-1 ; 1]. The output values of the ANN of controller was limited to the interval [-1 ; 1] (the values greater as 1 or smaller as –1 became 1 or -1).These values are converted in values in the interval [0 ; 255] ($V_{d8bv}$), which represents the universe of discourse of the *U* variable of the neural controller and of the inverse model.

In Figure 4 is presented the behavior of the ALCS. The desired illuminance has the value $E_{desired} = 100\ lx_{d8bv}$. The daylight trajectory [4] presents fast changes. The measured illuminance has an oscillatory behavior.

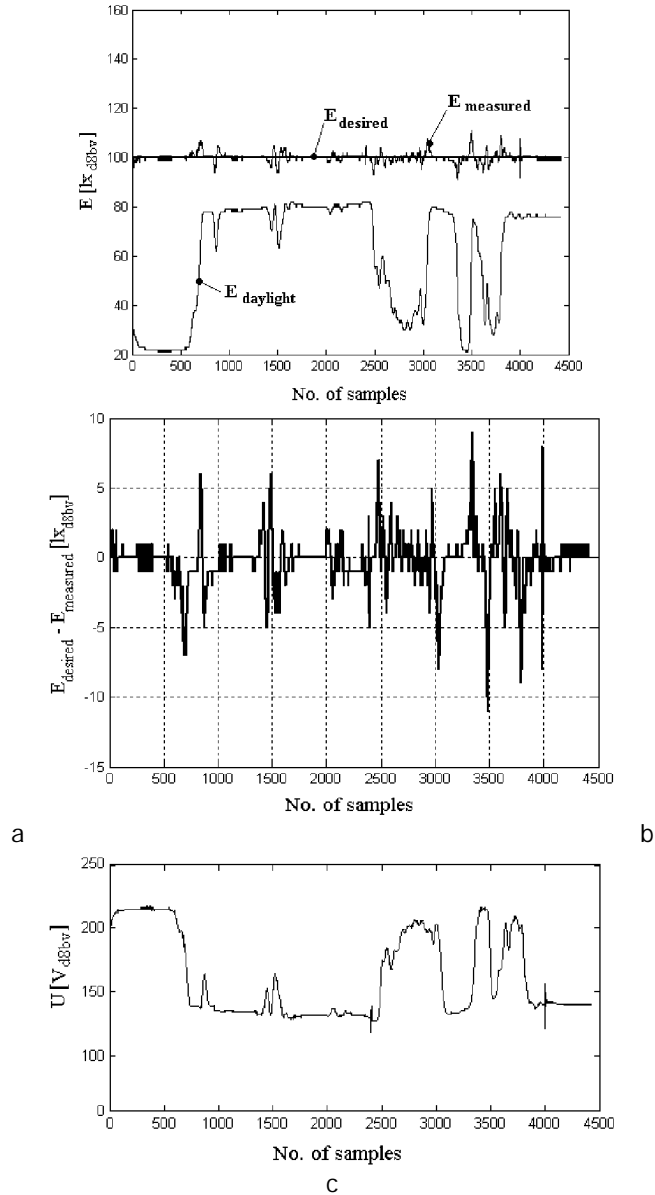

a	b

c

**Figure 4** Behavior of ALCS (learning rate $\gamma = 0,15$): (a) trajectories of illuminance: measured illuminance, desired illuminance and daylight illuminance; (b) error control trajectory; (c) control action (command) trajectory

Experimentally [5], a variation of measured illuminance in the interval [93;107] $lx_{d8bv}$ was not perceived by the human user. Analyzing the graphic from Figure 4b the steady-state error control has values in the interval [-11; 9] $lx_{d8bv}$. The extreme values of this interval are met rarely. The majority values for error control are in the interval [-5; 5] $lx_{d8bv}$. So, from the human eye perception, the human user of the ALCS does not perceive these oscillations.

**4 CONCLUSIONS**

The proposed structure used to control the lighting process, need an inverse model of the process. The mathematical model of the process is unknown. To solve the problem, an artificial neural network was trained on-line to reproduce the inverse model of the process. The controller was implemented with an artificial neural network too. In this way, the designer of the control scheme does not need any a priori information about the model of the lighting process. The control quality is not influenced by the precision of the experimental model of process that was used in [4, 5]. This control structure may be implemented on microprocessors because the structures of ANNs are not complicated, because the number of inputs and the number of neurons are small.

**REFERENCES**

1. Brown, M., Harris, C., "Neurofuzzy Adaptive Modeling and Control", Prentice Hall International (UK) Limited, 1994
2. Demuth, H., Beale, M., Hagan, M., "Neural Network Toolbox User's Guide", The MathWorks,Inc., 2007
3. Grif, H., Şt., Gyorgy, K., Gligor, A., Bucur, D., "Ways To Improve The Behavior Of A Neural Automatic Daylight Control System", Buletinul Ştiinţific al Universităţii "Petru Maior" din Târgu Mureş, Vol. XVII, 2004, p. 175-182
4. GRIF, H., Şt., Gligor, A., Bucur, D., "Fluorescent Daylight Control System based on B-spline Like Network whith Gaussian-Type Basis Functions", The 4[th] International Conference ILUMINAT 2007, Cluj-Napoca, 2007, pp. 16-1÷16-6;
5. Grif, H., Şt., Pop Mihaela, "Fluorescent daylight control system based on neural controller", INGINERIA ILUMINATULUI, Vol. 9, No. 19, 2007, pp. 14-22
6. Liang, X., Chen, R.,C., Yang, J., "An architecture-adaptive neural network online control system", Neural Comput & Applic (2008), 17:413–423 dr. Horaţiu Ştefan GRIF, lecturer
"Petru Maior" University from Tg. Mureş
1, N. Iorga St., RO-540088 Targu-Mures, Romania
Ph.:+40.766.644815
Fax:+40.265.236213
e-mail: hgrif@engineering.upm.ro, ghoratiu2000@yahoo.com